\documentclass{article}
\usepackage{spconf,amsmath,graphicx}
\usepackage{float}
\usepackage{booktabs}
\usepackage{cite}

\title{Dual stream computer-generated image detection network based on channel joint and softpool}
\name{Ziyi Xi,Hao Lin,Weiqi Luo*}
\address{School of Computer Science and Engineering, Sun Yat-sen University, GuangZhou  510006, China;}
\begin{document}
%
\maketitle
\begin{abstract}
With the development of computer graphics technology, the images synthesized by computer software become more and more closer to the photographs. While computer graphics technology brings us a grand visual feast in the field of games and movies, it may also be utilized by someone with bad intentions to guide public opinions and cause political crisis or social unrest. Therefore, how to distinguish the computer-generated graphics (CG) from the photographs (PG) has become an important topic in the field of digital image forensics. This paper proposes a dual stream convolutional neural network framework based on channel joint and softpool. The proposed network architecture includes a residual module for extracting image noise information and a joint channel information extraction module for capturing the shallow semantic information of image. In addition, we also design a residual structure to enhance feature extraction and reduce the loss of information in residual flow. The joint channel information extraction module can obtain the shallow semantic information of the input image which can be used as the information supplement block of the residual module. The whole network uses SoftPool to reduce the information loss of down-sampling for image. Finally, we fuse the two flows to get the classification results. Experiments on SPL2018 and DsTok show that the proposed method outperforms existing methods, especially on the DsTok dataset. For example, the performance of our model surpasses the state-of-the-art Quan\cite{quan2020learn} by a large margin of 3\%. \emph{The source code is accessible on github.} \footnote{https://github.com/zoie-ui/CG-Detection} 
\end{abstract}
\section{Introduction}
\label{sec:intro}
CG is the abbreviation of computer-generated graphics, which refers to the virtual but visually resonable images generated by computer software. PG is the abbreviation of Photographs, which means the real images taken by cameras. In recent years, CG technology has been widely used in the fields of games and movies. In this process, a large number of image processing tools were born, such as Vray, corona, Enscape and lumion. So people without professional knowledge can generate CG easily. As shown in Fig.1, it is difficult to distinguish between CG and PG by naked eyes. In addition, some studies have shown that although people's recognition rate will increase after receiving targeted training, they still can't effectively distinguish CG and PG, especially when they have doubts about a picture, they tend to identify it as a natural image \cite{holmes2016assessing}. This has raised people's concerns about safety, because these advanced CG technologies may be used by criminals to create realistic CG to confuse the public. Therefore, the study of CG detection technology has important academic significance and practical application value.

\begin{figure}[t]
\begin{minipage}[b]{0.48\linewidth}
  \centering
  \centerline{\includegraphics[width=4.0cm]{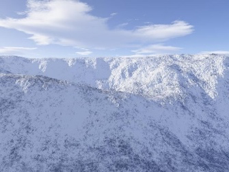}}
\end{minipage}
\begin{minipage}[b]{.48\linewidth}
  \centering
  \centerline{\includegraphics[width=4.0cm]{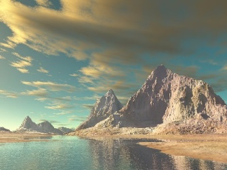}}
\end{minipage}
\centerline{(a) CG images}\medskip
\hfill
\end{figure}
\begin{figure}[t]

\begin{minipage}[b]{0.48\linewidth}
  \centering
  \centerline{\includegraphics[width=4.0cm]{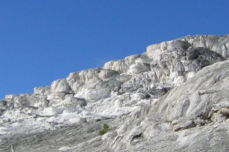}}
\end{minipage}
\begin{minipage}[b]{.48\linewidth}
  \centering
  \centerline{\includegraphics[width=4.0cm]{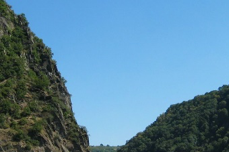}}
\end{minipage}
\centerline{(b) PG images}\medskip
\hfill
\caption{The visual comparison diagram of CG and PG, image examples from SPL2018.} \label{fig:eg}
\end{figure}

In this paper, we propose a novel dual stream convolutional neural network for CG detection. The proposed network is composed of the residual extraction module which learns the noise of image and the joint channel information extraction module which extracts the shallow semantic information of image.  The main contributions of our work can be summarized as follows:\par
\begin{itemize}
\item we propose a dual stream framework based on channel joint and softpool to solve the CG detection problem. The network consists of two main modules, namely, the residual extraction module and the channel joint feature extraction module. By fusing the features extracted from the two modules, we can achieve the best detection performance on SPL2018 and DsTok.
\item In the residual extraction module, we adopt a special residual structure, which can effectively enhance the learning of residual information.
\item we first introduce the SoftPool into the CG detection field, which reduces the information loss caused by downsampling.
\end{itemize}

\section{RELATED WORKS}
\label{sec:format}
Existing methods for CG detection can be generally divided into two categories——hand-crafted feature-based methods and deep learning-based methods. For traditional hand-crafted methods, It usually depends on statistics or internal characteristics difference of CG and PG, and demands people to design an efficient algorithm to extract features and make decisions between them. A simple strategy is to find a category sensitive scalar feature and select an appropriate classification threshold. The deep learning-based methods usually directly utilize deep neural network to autonomously learn complex features to complete classification.\par
Rahmouni et al.\cite{rahmouni2017distinguishing} first used convolutional neural network to learn a group of filters for image preprocessing and trained a multi-layer perception to complete the classification tasks. Quan et al. \cite{quan2018distinguishing} convinced that the detection accuracy of the model is directly affected by the image sampling mode. Therefore, they used the maximum poisson disk sampling to complete the data enhancement, and then trained a CNN with seven layers, finally produced the prediction result through the simple majority voting principle. Yao et al.\cite{yao2018distinguishing} proposed using high pass filters to remove the low-frequency component of the image, in other words, focusing on observing the sensor noise and residual introduced by the digital camera. Therefore, they designed three high-pass filters according to prior knowledge. The cropped image is first filtered by high pass filters, and then is transmitted to CNN for further learning. Quan et al. \cite{quan2020learn} proposed an attention network based CNN with 10 layers, which integrated RGB and filtered RGB images, with a total of 6 channels as the input of network. Zhang et al. \cite{zhang2020distinguishing} proposed a module composed of stacking convolution layers to preprocess R, G and B channel, then concat them by channel and transmit them to the five layers convolution neural network for further learning to get the results. He et al. \cite{he2020detection} took the six-channel image obtained by the channel fusion of the Gaussian filter preprocessed image and the original image as the input, and then sent it to a four layers dual stream network with different scales only in the first convolution layer, and finally fused the two streams through a simple attention mechanism.\par
Rezende et al. \cite{de2017detecting} based on transfer learning, send an RGB image after gray processing to the fine-tuning resnet50\cite{he2016deep} which had been pre-trained on ImageNet, achieved a good precision. It is worth mentioning that their time for detecting an image is 1.02s, which has a great significance for practical application. Nguyen et al. \cite{nguyen2018modular} found that the semantic information of the image will gradually lose with the increase of the depth of the neural network. This will make the features tend to be homogeneous, so they took the outputs of the first three layers as the extracted image features and transmitted them to the pre-constructed feature conversion module,
and then trained the classifier to obtain the detection results. Yao et al. \cite{yao2022cgnet} also extracted the output feature maps of the first three layers of vgg19\cite{simonyan2014very} as an input of the three stream network separately. The innovation is that the convolution block attention module is introduced before stream fusion to enhance the feature representation ability.
He et al. \cite{he2018computer} proposed a method that CNN combined with recurrent neural network (RNN) to detect CG. 

\begin{figure*}[t]
\centering
\includegraphics[width=1\linewidth]{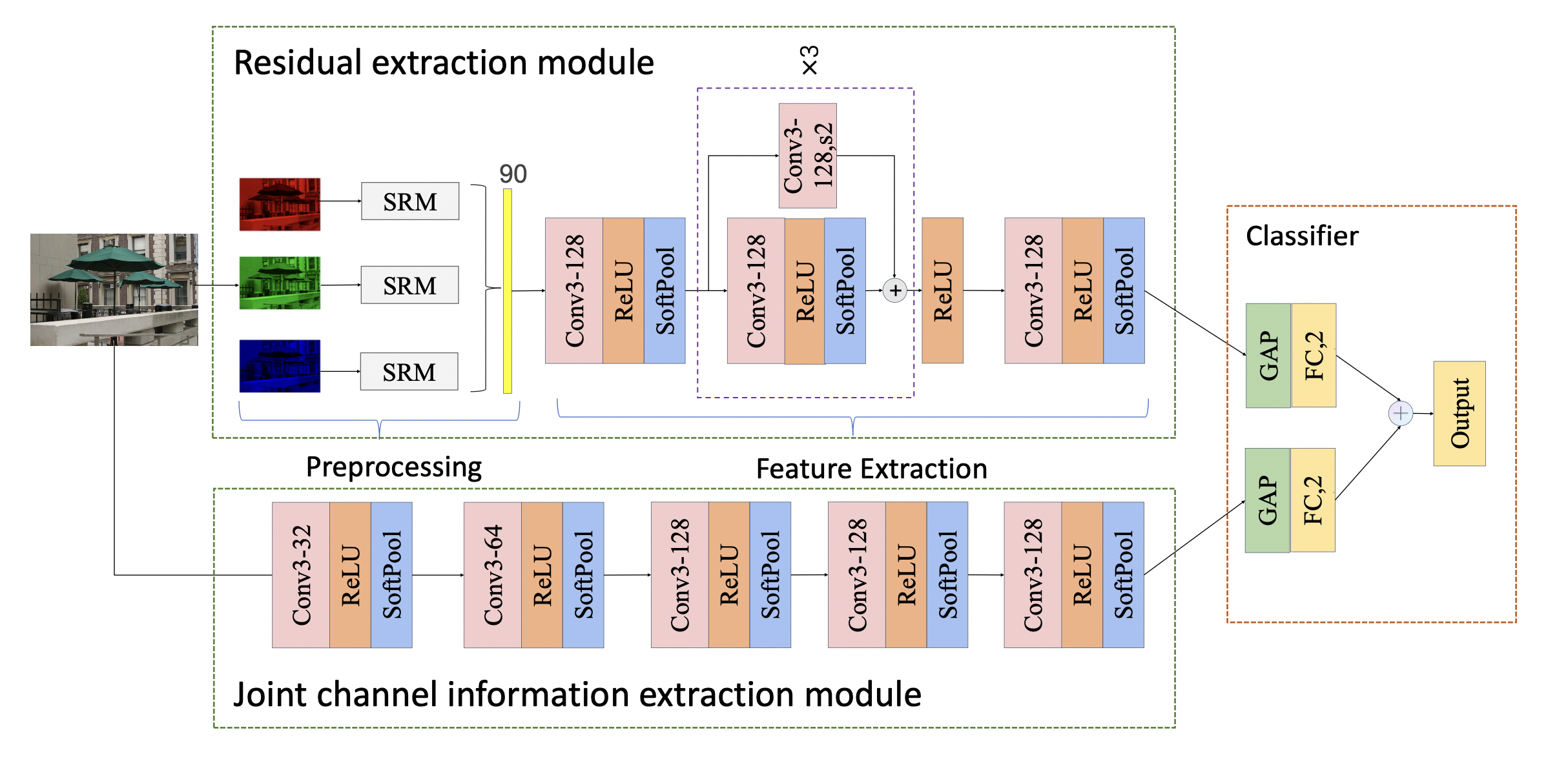}
\caption{The framework of the proposed network.} \label{fig:eg}
\end{figure*}

\section{Proposed Method}
\label{sec:pagestyle}
In this part, we will describe our network in detail. We propose a dual stream network to detect CG, which composed of residual stream extracting the residual information of image and joint channel stream extracting shallow semantic information. As shown in Figure 2, for the residual extraction module, we utilize the SRM to extract residual information of RGB color space through various channels, and then concat the feature maps by channel and further refined them by feature extraction network. Finally, we can obtain the 128 dimensional residual feature maps. In the joint channel information extraction module, we utilize the convolutional neural network to refine the original RGB image to gain the 128 dimensional shallow semantic feature maps. Besides, the whole network adopts SoftPool for downsampling, which is conducive to reduce the information loss in the process of network training. Finally, we merge the classification results of dual stream mentioned above to get the final output. Next we will introduce several module used in this network in detail.
\subsection{Preprocessing}
Inspired by Goljan M et al. \cite{goljan2014rich} and Quan et al. \cite{quan2020learn}, we utilize 30 SRM to filter the R, G and B channel respectively to extract the residual features, and then merge them by channel to obtain the 90 residual feature maps. It can strengthen the characterization of the relationship between local pixels in the same channel and get more complex statistical characteristics by concatnation, and also profits to enlarge the difference between PG and CG images.
\subsection{Feature extraction}
The network structure of the feature extraction module consists of five convolution layers, where each convolution layer is followed by a batch normalization, a ReLU activation function and a SoftPool. The purpose of the module is to fully learn the residual characteristics after fusion. There are three consecutive residual structures in the middle three layers.\par
The residual structure is mainly composed of two branches, as shown in Figure 3, where the previous layer input first passes through a convolution layer of size $3\times3$ with step 1, and then activated by a ReLU function, finally undergoes sampling in the SoftPool layer. There is a convolution layer of size $3\times3$ with step 2 in the branch. This, we can extract the residual information with the same scale. Finally, the results of two branches are fused by adding.\par
It is worth noting that we adopt SoftPool instead of MaxPool to down sample image. The basic functions of pooling layer include: reducing the amount of calculation, reducing model redundancy, preventing model over fitting and etc. Softpool \cite{stergiou2021refining} is a variant structure of pooling layer, it can enhance feature representation and retain the basic attributes of input. Specifically, SoftPool reduces the information loss caused by pooling while maintaining the basic functions of the pooling layer. Its calculation process is shown in Figure 4. 
For a region R of size $2\times2$, we first calculate the SoftMax value $w_{i} (i=1,2,3,4)$ of each pixel in the region, then multiply the SoftMax value and the original pixel value by elements, and accumulate the four values to obtain the pooled result  $\widetilde{\alpha}$. The specific formula as follows:\par
$$
\begin{array}{c}
w_{i}=\frac{e^{\alpha_{i}}}{\sum_{j \in R} e^{\alpha_{j}}}    \\
\\
\tilde{\alpha}=\sum_{j \in R} w_{j} \times \alpha_{j}
\end{array}
$$
where $\alpha_i$ represents the pixel value of the i-th pixel point, $w_i$ represents the weight corresponding to the i-th pixel.
\subsection{Joint channel information extraction module}
We adopt the original RGB image as the input of this module. It is also composed of five convolution layers. Similarly, each convolution layer of size $3 \times 3$ is followed by a batch normalization, a relu activation function and a SoftPool layer. Finally, we can also obtain the 128 dimensional shallow semantic information feature maps with the same size as the output of the residual extraction module.
Unlike natural images, which are limited by time, place and environment, CG contain many scenes that do not exist in reality. Therefore, the semantic content of image also contains important information for classification. After preprocessing, the original image content has been basically omitted. Therefore, the joint channel information extraction module can be the information supplement block of the residual module.

\begin{figure}[t]
\centering
\includegraphics[width=0.8\linewidth]{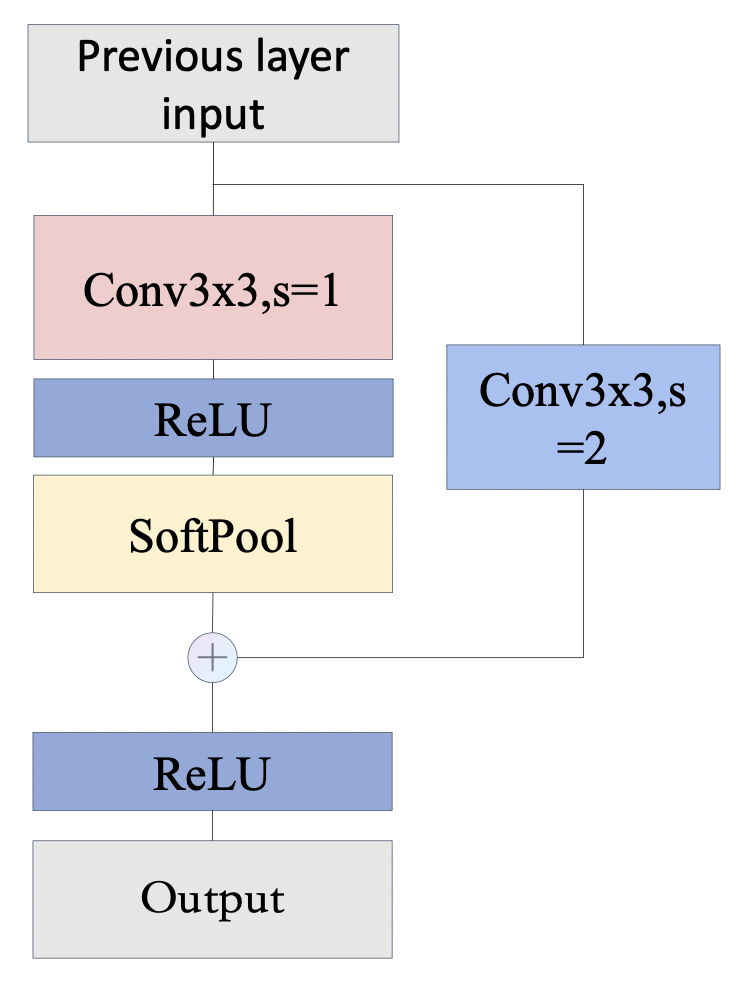}
\caption{The residual structure.} \label{fig:eg}
\end{figure}

\begin{figure}[t]
\centering
\includegraphics[width=1\linewidth]{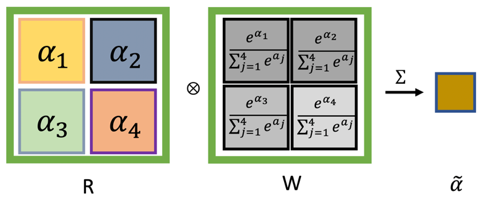}
\caption{The processing of SoftPool for a 2x2 region.} \label{fig:eg}
\end{figure}

\section{Experimental result}
\subsection{Datasets}
We conduct experiments on the SPL2018 and DsTok dataset. SPL2018 dataset was constructed by He et al.\cite{he2018computer}, which contained 6800 PG and 6800 CG. CG are collected from more than 50 rendering software, PG are taken by different types of camera under various environmental conditions, including indoor and outdoor. The range of image resolution is from $266\times199$ to $2048\times3200$, which is highly heterogeneous and difficult to detect. We divide it into training set, validation set and test set according to the ratio of 10:3:4.
DsTok dataset \cite{tokuda2013computer} contains 4850 PG and 4850 CG, and the image resolution are from $609\times603$ to $3507\times2737$. All CG and PG images in DsTok are collected from the Internet and have strong heterogeneity. We also divide it into three sets in the ratio of 3:1:1. In order to unify the input standard, we sampled the center region of each image with size $224\times224$ as the input of our model. 

\subsection{Experimental Settings}
Like most existing methods, we utilize accuracy (Acc) as our evaluation metrics, and its calculation formula can be expressed as follows:
$$
Acc = \frac{TP+TN}{P+N} \times 100 \%
$$
where P denotes the number of positive samples, in this paper represents the total number of natural images, N refers to the number of negative samples. TP and TN respectively refers to the number of positive samples and negative samples correctly classified.\par
Other experimental settings as follows: we use NVIDIA's Titan GPU to train our model in PyTorch deep learning framework. In the process of model training, we choose the cross entropy loss as our loss function, and exploit the SGD optimizer to optimize the model, where the size of mini-batch is 64. The initial learning rate is set as 1e-3, and is reductioned to 0.5 times of the original every 20 epochs. The weight decay rate is set as 1e-3. The total of training epochs is 120.
\subsection{Comparisons With Other State-of-the-art Methods}
We mainly compare with six existing advanced detection methods, and evaluated our model on DsTok and SPL2018 respectively. In order to make the experimental results more convincing, we randomly divided each dataset for three times and conducted fair test on each division. Finally, we adopt the average test result of three divisions as our final assessment value.\par
The experimental results are shown in Table 1. From the table we can see that our method has reached the best detection performance on both of the two datasets. Especially on DsTok, our model has been improved by 3\% than Quan\cite{quan2020learn}. On SPL2018, compared with the current advanced CG detection models such as Quan\cite{quan2020learn} and Yao\cite{yao2022cgnet}, the proposed method surpasses by 1.1\% and 0.4\% respectively. It is worth noting that Quan's method has a similar preprocessing procedure with ours, but they exploit MaxPool to down sample image in the whole network, which will cause the loss of information. In addition, Quan's method also ignores the relationship between residual information and image shallow semantic information.
\begin{table}[htbp]
  \setlength\tabcolsep{14pt}
  \centering
  \begin{tabular}{l@{}lc@{}lc@{}}
    \toprule
    \multicolumn{2}{c}{Method} & \multicolumn{1}{c}{DsTok} & \multicolumn{1}{c}{SPL2018} \\
    \midrule
    \multicolumn{2}{c}{Quan\cite{quan2018distinguishing}}      & \multicolumn{1}{c}{85.3\%} &  \multicolumn{1}{c}{89.4\%} \\
    \multicolumn{2}{c}{Yao\cite{yao2018distinguishing}}  & 
    \multicolumn{1}{c}{88.8\%} &  
    \multicolumn{1}{c}{89.8\%}  \\

    \multicolumn{2}{c}{He\cite{he2018computer}}     & \multicolumn{1}{c}{83.2\%} &  \multicolumn{1}{c}{88.0\%}   \\
    \multicolumn{2}{c}{Zhang\cite{zhang2020distinguishing}}      & \multicolumn{1}{c}{93.4\%} &  \multicolumn{1}{c}{92.8\%}    \\
    \multicolumn{2}{c}{Quan\cite{quan2020learn}} & 
    \multicolumn{1}{c}{93.9\%} &
    \multicolumn{1}{c}{92.8\%}  \\
    \multicolumn{2}{c}{Yao\cite{yao2022cgnet}} & 
    \multicolumn{1}{c}{92.1\%} &
    \multicolumn{1}{c}{93.5\%}  \\
    \multicolumn{2}{c}{Ours} & 
    \multicolumn{1}{c}{\textbf{96.9}\%} &
    \multicolumn{1}{c}{\textbf{93.9}\%}  \\
    \bottomrule
  \end{tabular}
  \caption{Comparisons with other methods.}
  \label{tab:example}
\end{table}
\subsection{Ablation Study}
In order to verify the rationality of our network structure, this section we mainly conducted the following ablation experiments: 1) Ablation of dual stream framework; 2) Performance analysis of different combinations of SRM residual filter cores; 3) The validity analysis of residual structure and the influence of its placement position on classification; 4) The impact of different pooling combinations on network performance.
\subsubsection{Ablation of dual stream framework}
The proposed network is a dual stream framework including residual extraction flow and joint channel information extraction flow. In order to explore the rationality of dual stream framework, we remove one of them(i.e. residual flow or joint channel flow) and compare them with the method in this paper. The experimental results are shown in Table 2.
From Table 2, we can see that on SPL2018, the result of the proposed method is 93.9\%, which increased 1\% and 2.9\% respectively than only exploit residual flow or only exploit joint channel flow. On DsTok dataset, the accuracy of single residual flow and single joint channel flow are 96.6\% and 84.5\% respectively. The detection result of the proposed network is 96.9\%, which is improved by 0.3\% and 12.4\% . It can be seen that the residual feature is an important feature to distinguish CG and PG images. It is also confirmed that the dual streams will get better results than any single flow of them.

\begin{table}[htbp]
    \centering
    \resizebox{0.4\textwidth}{!}{
    \begin{tabular}{l@{}c@{}c@{}}
        \toprule
         Model &  \text{\quad SPL2018\quad} & \text{\quad DsTok\quad} \\
        \midrule
         Only residual stream & 92.9\% & 96.6\% \\
         Only joint channel stream& 91.0\% & 84.5\% \\
         Ours & \textbf{93.9}\% & \textbf{96.9}\% \\
        \bottomrule
    \end{tabular}
    }
    \caption{Comparative results of single stream and the proposed dual stream.}
    \label{tab:my_label}
\end{table}

\begin{table}[htbp]\footnotesize
    \centering
    \resizebox{0.5\textwidth}{!}{
    \begin{tabular}{l@{}c@{}c@{}c@{}c@{}c@{}c@{}}
        \toprule
        Filter Set & \text{\quad $1^{st}$order} &  \text{\quad $2^{st}$order} & \text{\quad $3^{st}$order} & \text{\quad 3x3\quad} & \text{\quad 5x5 \quad} & \text{\quad Ours\quad} \\
        \midrule
        Acc & 94.6\% & 95.1\% &  95.7\% & 95.3\% & 95.5\%  & \textbf{96.9}\%  \\
        \bottomrule
    \end{tabular}
    }
    \caption{Comparative studies for different SRM filter combinations.}
    \label{tab:my_label}
\end{table}

\subsubsection{Different combinations of SRM residual filter cores}
In this paper, we exploit 30 residual filtering cores to filter each channel. This section we will explore the impact of using the combinations of different filtering cores on the final performance of the model. According to the division of filter cores in \cite{goljan2014rich}, we conduct five groups of ablation experiments, including first-order filter cores (8), second-order filter cores (4), third-order filter cores (8), $3\times3$ filter cores (17, including 12 filled first-order and second-order filter cores , 4 edge $3\times3$ filter cores and square $3\times3$ filter core) and $5\times5$ filter cores (13, including 8 filled third-order filter cores, 4 edge $5\times5$ filter cores and 1 square $5\times5$ filter core). Besides, except for the preprocessing part of the residual extraction module, the rest part of the network remains unchanged. In addition, the experiments are only conducted on the DsTok dataset. The experimental results are shown in Table3, where we can observe that using 30 residual filter cores (Ours) can achieve the best performance for CG detection, and the Acc can reach 96.9\% on DsTok.

\subsubsection{residual structure}
In the residual extraction module, we also designed a residual structure. In order to verify the effectiveness of this structure, we designed ablation experiments as follows: 
\begin{itemize}
    \item VA: Neither residual flow nor joint channel flow exploit residual structure in the middle three layers.
    \item VB: Residual flow does not use residual structure, and joint channel flow exploits the structure in the middle three layers. 
    \item VC: Both residual flow and joint channel flow exploit this structure in the middle three layers. 
\end{itemize}
Our method is that the residual flow uses the residual structure in the middle three layers, and the joint channel flow does not use the residual structure. Other layers remain unchanged, and the experimental results
see Table 4. Here we can see that the use of residual structure in the middle three layers of residual flow can greatly increase the detection performance on DsTok, which proves that the proposed residual structure can indeed enhance the feature learning of residual flow and improve the classification accuracy. One possible explanation is that the similar network structure makes the learning characteristics of the two flows tend to be homogeneous. Therefore, it is determined that the residual structure is ultimately used in the residual flow and not in the joint channel flow.\par
Next, we will continue to discuss the impact of the location setting of the residual structure on the residual flow. We have set up three groups of ablation experiments for residual flow as follows.
\begin{itemize}
    \item[(1)]Ours (3 layers): The first and the last layer of the network are ordinary convolution layer, and the layer 2-4 are residual structure.
    \item[(2)]4 layers: Layer 2-5 are residual structure, and layer 1 is ordinary convolution layer;
    \item[(3)]5 layers: The whole layers utilize the residual structure.
\end{itemize}
   The experimental results are shown in Table 5. From the table, we can see that on SPL2018, the result of exploiting 4 layers is the best, with an accuracy of 93.1\%, which has a weak advantage over using 3 layers and 5 layers. However, on DsTok, using 3 layer outperforms the other methods and has fewer network parameters. Therefore, we choose exploiting the residual structure in the middle three layers of residual flow.
\begin{table}[htbp]
  \setlength\tabcolsep{10pt}
  \centering
  \begin{tabular}{l@{}lc@{}lc@{}}
    \toprule
    \multicolumn{2}{c}{Structure} & \multicolumn{1}{c}{SPL2018} & \multicolumn{1}{c}{DsTok} \\
    \midrule
    \multicolumn{2}{c}{VA}      & \multicolumn{1}{c}{\textbf{94.1}\%} &  \multicolumn{1}{c}{94.3\%} \\

    \multicolumn{2}{c}{VB}     & \multicolumn{1}{c}{93.8\%} &  \multicolumn{1}{c}{94.4\%}   \\
    
    \multicolumn{2}{c}{VC}      & \multicolumn{1}{c}{93.5\%} &  \multicolumn{1}{c}{96.4\%}    \\
  
    \multicolumn{2}{c}{Ours} & 
    \multicolumn{1}{c}{93.9\%} &
    \multicolumn{1}{c}{\textbf{96.9}\%}  \\
    \bottomrule
  \end{tabular}
  \caption{Comparative studies for different residual structure.}
  \label{tab:example}
\end{table}
\begin{table}[htbp]
  \setlength\tabcolsep{10pt}
  \centering
  \begin{tabular}{l@{}lc@{}lc@{}}
    \toprule
    \multicolumn{2}{c}{Layer} & \multicolumn{1}{c}{SPL2018} & \multicolumn{1}{c}{DsTok} \\
    \midrule
    \multicolumn{2}{c}{4 layers}      & \multicolumn{1}{c}{\textbf{93.1}\%} &  \multicolumn{1}{c}{95.3\%} \\

    \multicolumn{2}{c}{5 layers}     & \multicolumn{1}{c}{93.0\%} &  \multicolumn{1}{c}{95.7\%}   \\
    
    \multicolumn{2}{c}{Ours(3 layers)}      & \multicolumn{1}{c}{92.9\%} &  \multicolumn{1}{c}{\textbf{96.6}\%}    \\
    \bottomrule
  \end{tabular}
  \caption{Comparative results for using different residual structures in the proposed model.}
  \label{tab:example}
\end{table}

\begin{table}[htbp]
  \setlength\tabcolsep{10pt}
  \centering
  \begin{tabular}{l@{}lc@{}lc@{}}
    \toprule
    \multicolumn{2}{c}{pooling combination} & \multicolumn{1}{c}{SPL2018} & \multicolumn{1}{c}{DsTok} \\
    \midrule
    \multicolumn{2}{c}{M1}      & \multicolumn{1}{c}{93.8\%} &  \multicolumn{1}{c}{96.1\%} \\

    \multicolumn{2}{c}{M2}     & \multicolumn{1}{c}{93.5\%} &  \multicolumn{1}{c}{95.7\%}   \\
    
    \multicolumn{2}{c}{M3}     & \multicolumn{1}{c}{93.7\%} &  \multicolumn{1}{c}{96.3\%}   \\
    
    \multicolumn{2}{c}{Ours}      & \multicolumn{1}{c}{\textbf{93.9\%}} &  \multicolumn{1}{c}{\textbf{96.9\%}}    \\
    \bottomrule
  \end{tabular}
  \caption{Comparative results for using different pooling combinations in the proposed model.}
  \label{tab:example}
\end{table}
\subsubsection{pooling combinations}
In this section, we will explore the effectiveness of using various combinations of pooling. We have set up four groups of experiments:
\begin{itemize}
    \item M1:Both residual flow and joint channel flow use the MaxPool.
    \item M2:Residual flow utilizes the SoftPool, and joint channel flow utilizes the MaxPool.
    \item M3:Residual flow utilizes the MaxPool, and joint channel flow utilizes the SoftPool.
    \item Ours: Both residual flow and joint channel flow use the SoftPool.
\end{itemize}
The experimental results are shown in Table 6, where we can see that if we utilize the other combinations of pooling, the detection accuracy decreased in varying degrees on both of two datasets. This indicates that SoftPool profits to CG detection.

\section{Summary}
In this paper, we propose a dual stream convolution neural network for CG detection, which includes a residual extraction flow for learning the noise of image and a joint channel information extraction flow for learning shallow semantic information. In addition, we also designed an effective residual structure, which can reduce the loss of information caused by the pooling layer and it also profits to improve the detection ability of the model. We have evaluated several related works on two mainstream CG detection datasets, namely DsTok and SPL2018. A large number of comparative experiments show that the proposed method has achieved the best detection performance at present. In addition, we also designed a series of ablation experiments to verify the rationality of the proposed network structure. Although the model in this paper has achieved the current optimal detection effect on the current two mainstream datasets, there is still room for further improving. Next we will resort to explore the difference of CG and PG in spatial domain and frequency domain. At the same time, we plan to introduce attention and other mechanisms, and organically integrate these features into the existing network framework to enrich the extracted features and further improve the detection performance of the model.

\end{document}